\documentclass[runningheads]{llncs}
\usepackage{algpseudocode}
\usepackage{amsmath,amssymb,amsfonts}
\usepackage{array}
\usepackage{booktabs}       % professional-quality tables
\usepackage{enumitem}
\usepackage[T1]{fontenc}    % use 8-bit T1 fonts\
\usepackage{graphicx}
\usepackage{hyperref}       % hyperlinks
\usepackage[utf8]{inputenc} % allow utf-8 input
\usepackage{microtype}      % microtypography
\usepackage{multirow}
\usepackage[numbers,sort&compress]{natbib}
\usepackage{nicefrac}       % compact symbols for 1/2, etc.
\usepackage{siunitx}
\usepackage{subcaption}
\usepackage{textcomp}
\usepackage{url}            % simple URL typesetting
\usepackage[dvipsnames]{xcolor}
\usepackage{nicefrac}

\usepackage{listings}
\usepackage{color}
\usepackage{pifont}

\newcommand\crule[3][black]{\textcolor{#1}{\rule{#2}{#3}}}

\newcolumntype{L}[1]{>{\raggedright\let\newline\\\arraybackslash\hspace{0pt}}m{#1}}
\newcolumntype{C}[1]{>{\centering\let\newline\\\arraybackslash\hspace{0pt}}m{#1}}
\newcolumntype{R}[1]{>{\raggedleft\let\newline\\\arraybackslash\hspace{0pt}}m{#1}}

\usepackage{tikz}

\begin{document}
\title{Random Bundle: Brain Metastases Segmentation Ensembling through Annotation Randomization}
\titlerunning{Random Bundle Ensemble}

\author{Darvin Yi\inst{1}\thanks{Corresponding Author: darvinyi@stanford.edu} \and
Endre Gr{\o}vik\inst{2} \and
Michael Iv\inst{3} \and
Elizabeth Tong\inst{3} \and
Greg Zaharchuk\inst{3} \and
Daniel Rubin\inst{1,3}}
\authorrunning{D. Yi et al.}
\institute{Dept. of Biomedical Data Science at Stanford University, Stanford, CA 94305 USA\\
%\email{\{darvinyi,rubin\}@stanford.edu}
\and
Dept. for Diagnostic Physics at Oslo University Hospital, Oslo, Norway\\
%\email{endre.grovik@mn.uio.no}
\and
Dept. of Radiology at Stanford University\\
%\email{\{miv,etong,gregz,rubin\}@stanford.edu}
}

%\author{First Author\inst{1}\thanks{Corresponding Author: fauthor@university.edu} \and
%Second Author\inst{2} \and
%Third Author\inst{3} \and
%Fourth Author\inst{1} \and
%Fifth Author\inst{2} \and
%Sixth Author\inst{3}}
%\institute{Dept. of Science at University, City, State Code Country\\
%%\email{\{fauthor,fauthor\}@university.edu}
%\and
%Dept. of Science at University, City, State Code Country\\
%%\email{\{sauthor,fauthor\}@university.edu}
%\and
%Dept. of Science at University, City, State Code Country\\
%%\email{\{tauthor,sauthor\}@university.edu}
%}

\maketitle              % typeset the header of the contribution
\renewcommand*{\thefootnote}{\fnsymbol{footnote}}
\setcounter{footnote}{3}
\begin{abstract} %15-250 words

We introduce a novel deep learning segmentation network ensembling method, Random Bundle (RB), that improves performance for brain metastases segmentation.  We create our ensemble by training each network on our dataset with 50\% of our annotated lesions censored out.  We also apply a lopsided bootstrap loss to recover performance after inducing an \emph{in silico} 50\% false negative rate and make our networks more sensitive.  We improve our network detection of lesions's mAP value by 39\% and more than triple the sensitivity at 80\% precision.  We also show slight improvements in segmentation quality through DICE score.  Further, RB ensembling improves performance over baseline by a larger margin than a variety of popular ensembling strategies.  Finally, we show that RB ensembling is computationally efficient by comparing its performance to a single network when both systems are constrained to have the same compute.

\end{abstract}

\section{Introduction}

The rise of deep learning (DL) has created many novel algorithms for segmentation, from one of the first uses of a deconv (or convolutional transpose) layer \cite{long2015fully} or hourglass networks \cite{ronneberger2015u,badrinarayanan2017segnet} to more hierarchical spatially aware networks \cite{zhao2017pyramid}.  Recently, further advances have enabled networks to increase retinal field through the use of atrous (dilated) convolutions \cite{chen2017rethinking}.  These computer vision (CV) advances have translated incredibly well to medical image informatics, especially through carefully curated datasets \cite{bakas2018identifying,pedrosa2019lndb,setio2017validation}%,armato2011lung}
that have led to the design of many medical image-specific network architectures \cite{milletari2016v,kamnitsas2016deepmedic,zhou2018unet++}.  A more complete summary for brain tumor segmentation can be found with I{\c{s}}{\i}n \emph{et al.} \cite{icsin2016review} and a more recent review of DL for medical imaging with Hesamian \emph{et al.} \cite{hesamian2019deep}.

In our work, we study the benefits of ensembling several (up to $n=30$) networks trained on the same task of brain metastases segmentation.%  Ensembling has long been a part of deep learning, including ensembling for classification \cite{simonyan2014very} on the ImageNet task.
Medical image networks often face the challenge of creating an ensemble of networks with different tasks, such as training and fusing three different view-specific networks \cite{lyksborg2015ensemble}.  Other groups have ensembled several classification networks for the purpose of segmentation\cite{jia2018atlas}.  However, previous work indicates that taking an ensemble of several networks trained on the same task improve results modestly, with gains of 1-3\% \cite{li2018fully,lahiri2016deep}.  The most impressive work of Kamnitsas \emph{et al.} \cite{kamnitsas2016deepmedic} shows a statistically well motivated approach to ensembling networks, but deals mainly with ensembling three different networks of different architectures that have all been individually optimized.  In our work, we will study the idea of network ensembles in the same vein as tree bagging, finding a DL segmentation analogue to Random Forest.

We view the following as our main novel contributions:
\begin{enumerate}
    \item to the best of our knowledge, the first study of DL network ensembles for brain lesion segmentation
    \item a thorough comparison of the common ensembling techniques
    \item creation of a novel ensembling methodology named Random Bundle that builds off Yi \emph{et al.}'s false negative robustness work \cite{yi2020brain}
    \item to the best of our knowledge, the first compute-normalized study of ensemble performance for medical image segmentation
\end{enumerate}

\section{Data}

We use a metastases segmentation dataset \cite{grovik2020deep}, consisting of brain MR images.  Expert annotations for metastasis lesions were created/verified by two neuroradiologists with a combined experience of 13 years.  (A subset of the annotations were initially labeled by a fourth-year medical student and then additionally edited/verified.)  Labels were created using the polygon tool on a professional version of OsiriX.  There is no overlap in neuroradiologist annotations, which precludes any statistics on radiologist variability.  This method maximizes the number of labels but provides no information on the quality of our target annotations.

The dataset contains 156 patients, split 67\%-33\% for training-testing.  The testing set was created with 17 patients from each subgroup of lesion count: 1-3 lesions, 4-10 lesions, and more than 10 lesions.  Each patient has four MR pulse sequences that were co-registered with nordicICE (Nordic Neuro Lab, Bergen, Norway).  The four sequences comprise a pre- and post-contrast T1-weighted 3D fast spin echo (CUBE), a post-Gd T1-weighted 3D axial inversion recovery prepped fast spoiled gradient-echo (BRAVO), and a 3D fluid-attenuated inversion recovery (FLAIR).  Annotations were done on the BRAVO series, and all other pulse sequence images were spatially registered to the BRAVO series.

\section{Methods}

We test different ensembling techniques, including our novel ``Random Bundle''\footnote{in homage to Random Forest, but for neural networks} (RB) method.  We also include an additional ``wide-res-net'' experiment that will show that RB ensembling is an efficient use of compute.

%\subsection{Review of Bootstrap Loss}

\textbf{The Lopsided Bootstrap Loss} \ \ \ \ \ Based on previous work \cite{reed2014training,yi2020brain}, the ``lopsided bootstrap loss'' (Eq. \ref{eq:lopsided}) can be used to train a segmentation network on labeled data with a very high false negative rate (50\%) of annotations.  Training with the bootstrap loss improved network sensitivity, which we hope to leverage together with ensembling in this work.

\begin{equation}\label{eq:lopsided}%\vspace{-1mm}
\mathcal{L} (Y,\hat{Y}) = \left\{\begin{array}{lc}
\beta * CE(Y,\hat{Y})+ (1-\beta)*CE(\text{argmax}(\hat{Y}),\hat{Y}))&\text{if } Y == 0\\
\alpha * CE(Y,\hat{Y})&\text{if } Y == 1
\end{array}
\right.
\end{equation}

In Eq. \ref{eq:lopsided}, we introduce the hyperparameters $\alpha > 1$, which describes upweighting of positive cases, and $\beta \in (0,1]$, which describes the percentage of loss ascribed to the bootstrap factor.  For example, 1 corresponds to normal cross entropy only, while 0.5 gives equal 50\%-50\% split between normal and bootstrap cross entropy.

%\subsection{Ensembling Methods}\label{sec:ensembling}

\textbf{Ensembling Different Hyperparameters and Initializations} \ \ \ \ \ The most basic form of network ensembling is to train several different networks, each with different hyperparameters and weight initializations.  This proves to be the easiest to implement, mainly because such networks will naturally be trained during hyperparameter search.  Thus, saving different checkpoints of the model between hyperparameter search runs can give the basis for this naive ensembling approach.

\begin{figure}[htb!]
	\centering
	\begin{subfigure}[h]{0.23\textheight}
		\includegraphics[height=0.23\textheight]{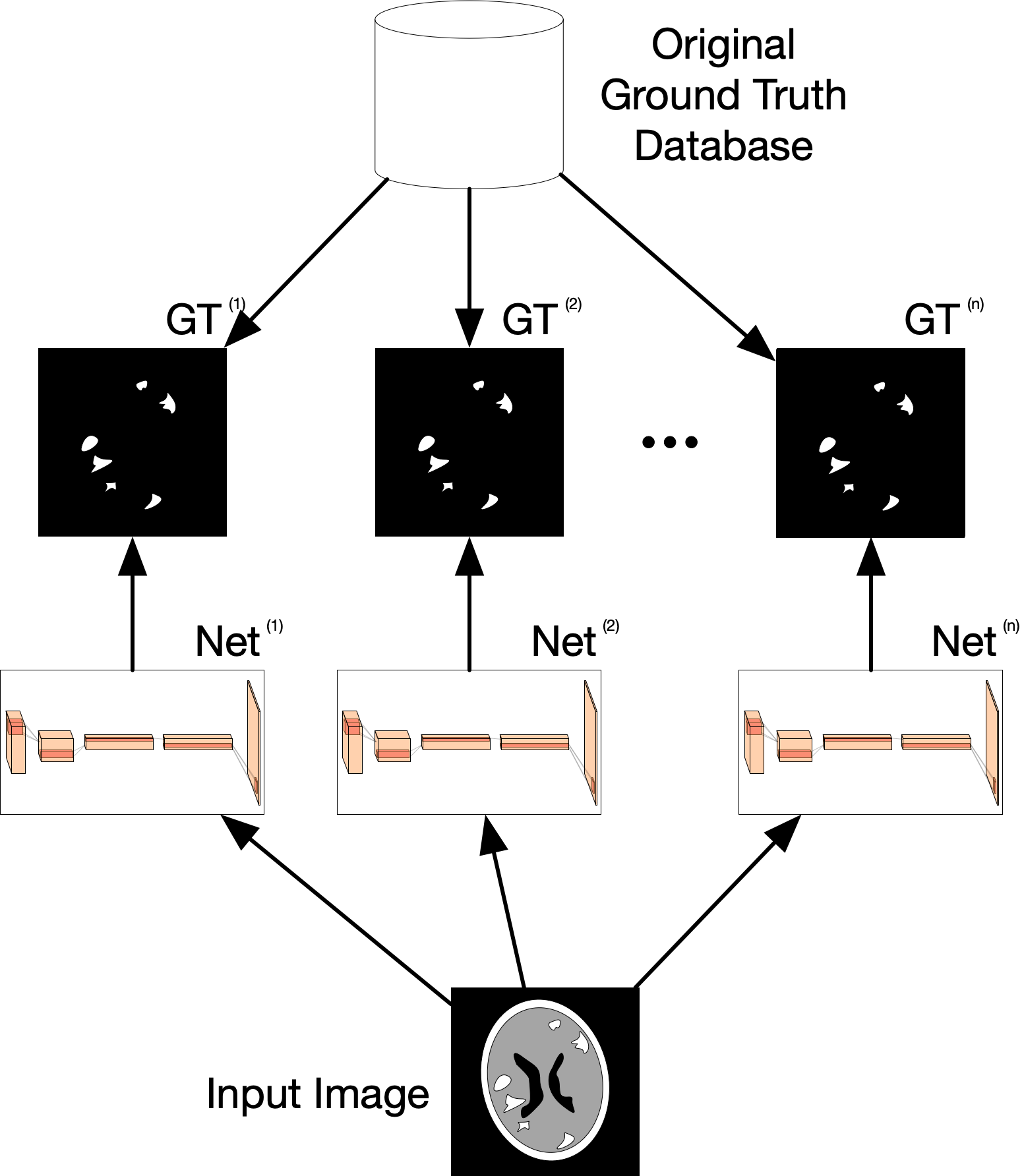}
		\caption{Basic Ensembling Concept}
	\end{subfigure}
	~
	\begin{subfigure}[h]{0.23\textheight}
		\includegraphics[height=0.23\textheight]{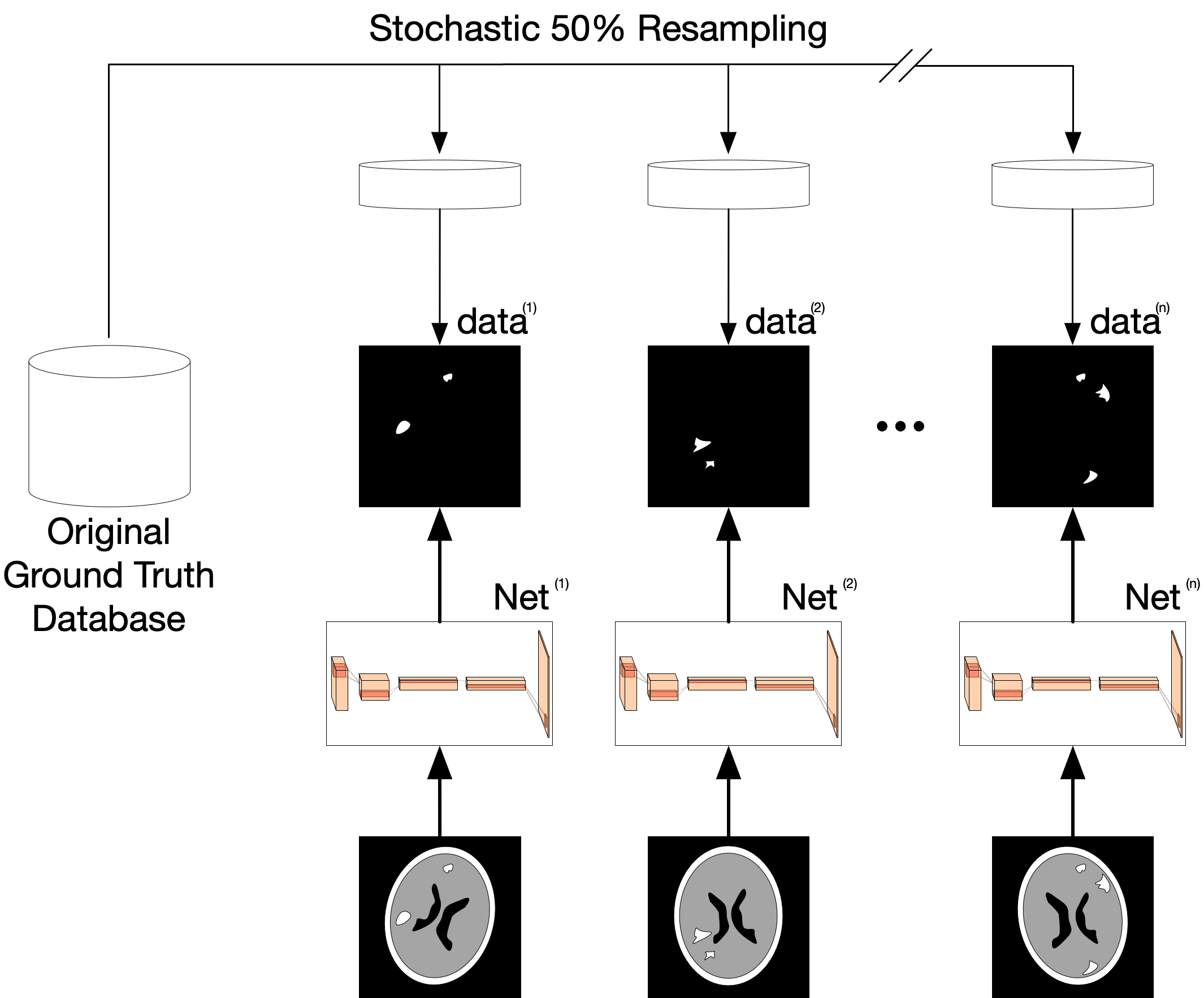}
		\caption{Random Data Subsampling}
	\end{subfigure}
	\caption{\textbf{Known Ensembling Methods.} \small{(a) Basic concept of ensembling.  (b) Ensembling by training on randomly subsampled patient cohorts.}}
	\label{fig:ensembling}
\end{figure}

\textbf{Ensembling Different Network Architectures} \ \ \ \ \ This technique ensembles different network architectures, all trained on the same task.  For this segmentation task, we use 6 different architectures important in semantic segmentation: FCN \cite{long2015fully}, SegNet \cite{badrinarayanan2017segnet}, U-Net \cite{ronneberger2015u}, FPN \cite{lin2017feature}, PSP Net \cite{he2015spatial}, and DeepLabv3 \cite{chen2017rethinking}.

\textbf{Ensembling Networks Trained on Different Patient Cohorts} \ \ \ \ \  This method involves training $n$ different networks, each on a random sampling of half of the available patients (Fig. \ref{fig:ensembling}b).  Note that each patient has probability $0.5^n$ of being omitted from all sampled datasets, or 9.3e-10 given our choice of $n = 30$.  With 100 patients, the total chance we will not see any single patient is 9.3e-8, a trivial chance.

\begin{figure}[htb!]
	\centering
	\begin{subfigure}[h]{0.25\textheight}
		\includegraphics[width=\textwidth]{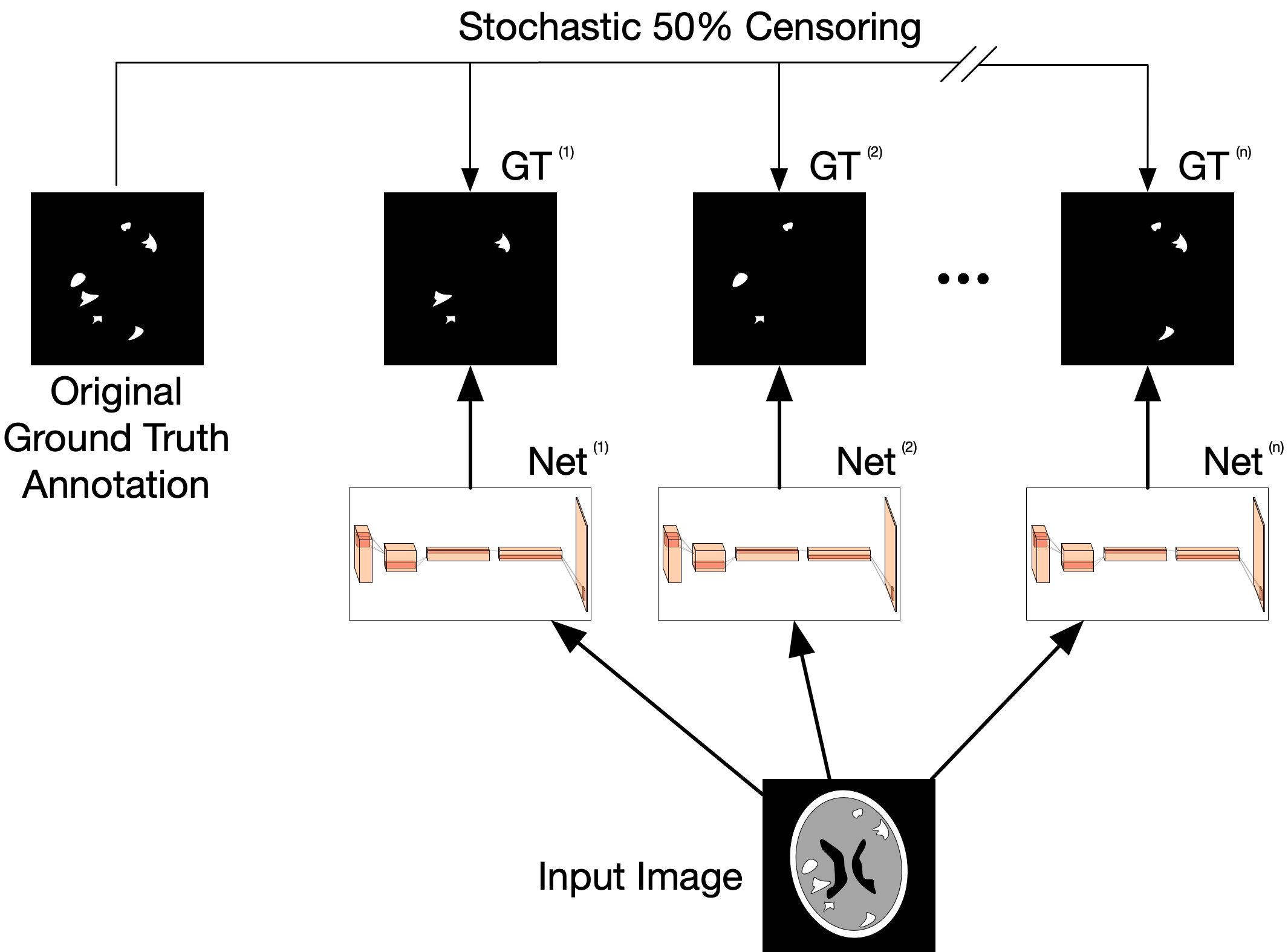}
		\caption{RB at Train Time}
	\end{subfigure}
	~
	\begin{subfigure}[h]{0.25\textheight}
		\includegraphics[width=\textwidth]{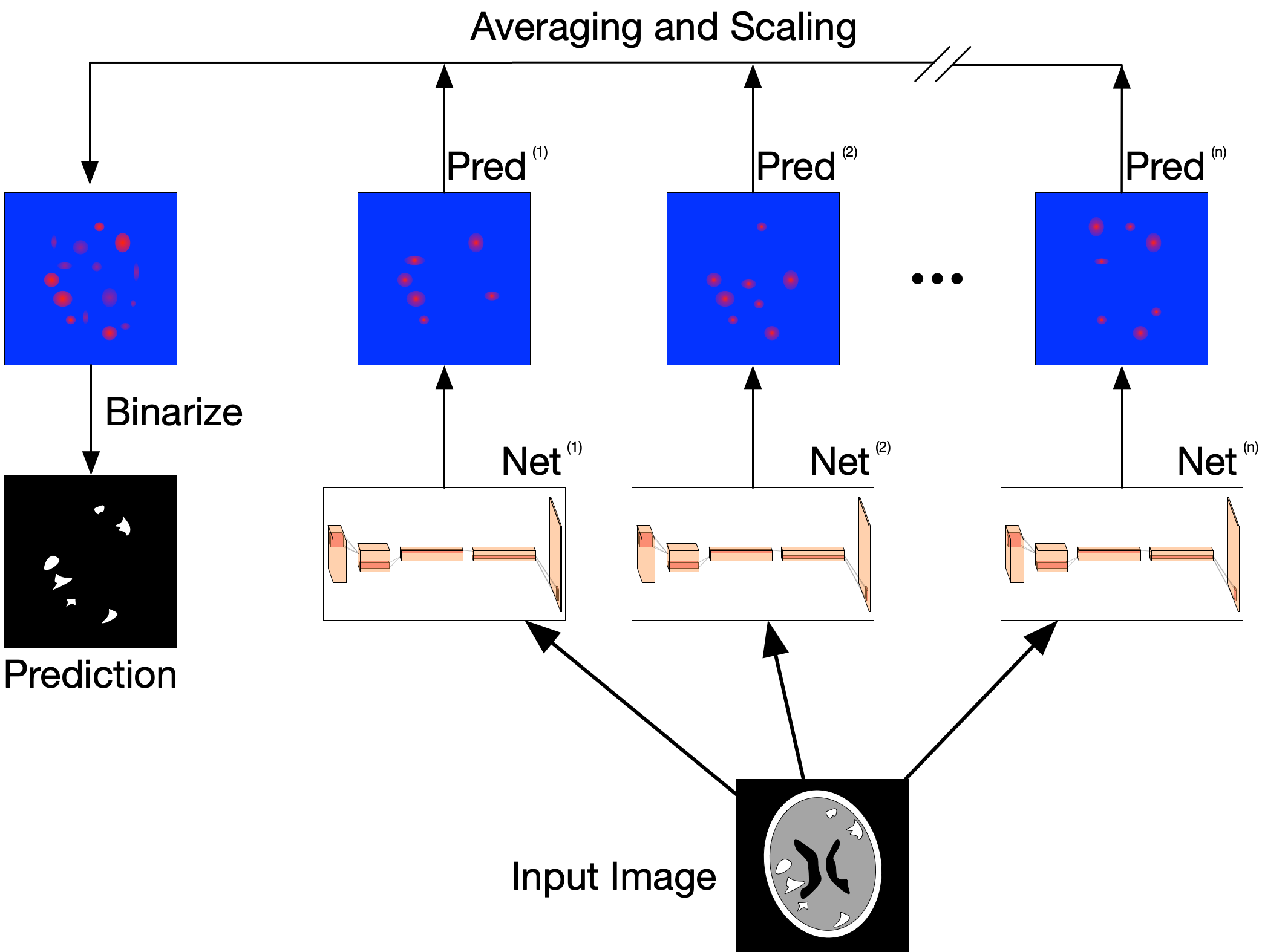}
		\caption{RB at Inference Time}
	\end{subfigure}
	\caption{\textbf{Proposed ``Random Bundle'' Ensembling.} \small{(a) Training Schema.  Each Network in the ensemble is trained on all the patients, but with a random 50\% of lesions censored out across all patients. (b) RB at Inference Time.}}
	\label{fig:RB}
\end{figure}

\textbf{Ensembling Networks Trained on Different Lesion Censoring} \ \ \ \ \  This is our main novel contribution to ensembling, named ``Random Bundle'' (in homage to Random Forest).  As above, we train each network of our ensemble on a different sampling of the dataset; however, dataset will be stochastically sampled by lesions, not patients (Fig. \ref{fig:RB}a).  We censor lesions with the same 50\% chance, so any single lesion is omitted with the same probability of 9.3e-10.  However, with 1780 lesions, our total probability of not seeing any single lesion is a higher 1.7e-6.  Nonetheless, this probability is still small enough to conduct this experiment without a large fall out.

%\subsection{Wide Res-Net Experiment}

\textbf{Wide Res-Net Experiments} \ \ \ \ \ We use a wide res-net (WRN) \cite{zagoruyko2016wide} to explore compute use, modifying the original architecture by replacing the end-layers (average pool followed by fully connected neural network) with a single convolutional transpose layer back to original resolution.  Compute increases quadratically with width parameter $k$.  Therefore, we can compare an ensemble of 16 WRNs with $k=\nicefrac12$ using our RB ensemble to a single WRN with $k=2$ (twice that of the original res-net).

%\subsection{Implementation Details}

\textbf{Implementation Details}  \ \ \ \ \ We follow published methods \cite{yi2020brain,grovik2020deep} to train our brain metastases segmentation network.  We use the same 2.5D network structure with 5 z-slices (centered on the prediction slice) of all four pulse sequence images.  These 20 slices were stacked in the color channel and passed into the network as a single $256 \time 256 \times 20$ tensor.

In our hyperparameter ensembling experiment, we test a range of learning rates, $L_2$ regularization parameters, and optimizers:\\
\noindent $\text{Learning Rate} \in \{\num{3e-2},\num{1e-2},\num{3e-3},\num{1e-3},\num{3e-4}\}$,\\
\noindent $\lambda^{(L_2)} \in \{\num{1e-2},\num{1e-3}\}$, $\text{Optimizers} \in \{\text{Momentum}, \text{RMSProp},\text{Adam}\}$.
%\begin{align*}
%    \text{Learning Rate} &\in \{\num{3e-2},\num{1e-2},\num{3e-3},\num{1e-3},\num{3e-4}\}\\
%    \lambda^{(L_2)} &\in \{\num{1e-2}, \num{1e-3}\}\\
%    \text{Optimizers} &\in \{\text{SGD Momentum}, \text{RMSProp},\text{Adam}\}
%\end{align*}
Every combination of these three hyperparameters gives us a network in our eventual ensemble.  Using the validation set, we found that the best set $\{\text{LR}, L_2, \text{Opt}\}$ is $\{\num{3e-3},\num{1e-2},\text{Adam}\}$, which we use in training all subsequent networks.  (While further network performance could have been better optimized through additional hyperparameter selection, we opted to forgo this process due to the large number of experiments and trained networks.)  For a single network, we report the value of the best network (based on the validation set).  For an ensemble of 3 or 10 networks, we randomly chose the subset of networks to ensemble.

All code was written in Python with the Pytorch framework\cite{paszke2017automatic} and run on a commercial-grade workstation: intel i7 9750k CPU, two NVIDIA GTX 1080Ti GPU's, 64gB CPU-RAM.  Each single pass prediction on a DeepLabv3 Network would take less than 100ms per slice.  Other smaller networks (e.g. Original FPN) took less time.  Thus, with the upper-bound of slice-wise prediction at 100ms, a full 3D MR scan would take 30s for prediction, ensuring that our largest ensembles of 30 networks would take a forward pass time of 15min to run a full 3D MR scan through all networks and aggregate the predictions.

%\subsection{Metrics}\label{sec:metrics}

\textbf{Metrics}  \ \ \ \ \ We report the same metrics as previous brain metastases segmentation papers \cite{yi2019mri,yi2020brain}, except for one difference.  We continue to report the mean average precision (mAP) with respect to the detection of brain lesions.  As in previous work, 3D segmentation probability maps are binarized with probability threshold 10\% to calculate the centroid of each 3D connected component (CC); if the centroid is within 1mm of any expert annotation, we treat the 3D CC as a true positive annotation.  The CC's predicted probability is set to be the average probability of all voxels in the CC from the original segmentation map.  From this, we can create a precision-recall (PR) curve and calculate its 95\% confidence intervals (CIs) \cite{hanley1983method}.  However, we report the sensitivity at 80\% precision (rather than maximum sensitivity), as this better demonstrates the increase in performance due to ensembling and represents the full PR curve.  Finally, we also report the 3D DICE scores for the true positive CCs at 80\% precision.

\section{Results}

%\subsection{Comparing Ensembling Techniques}

\begin{figure}[htb!]
	\centering
	\includegraphics[width=0.8\textwidth]{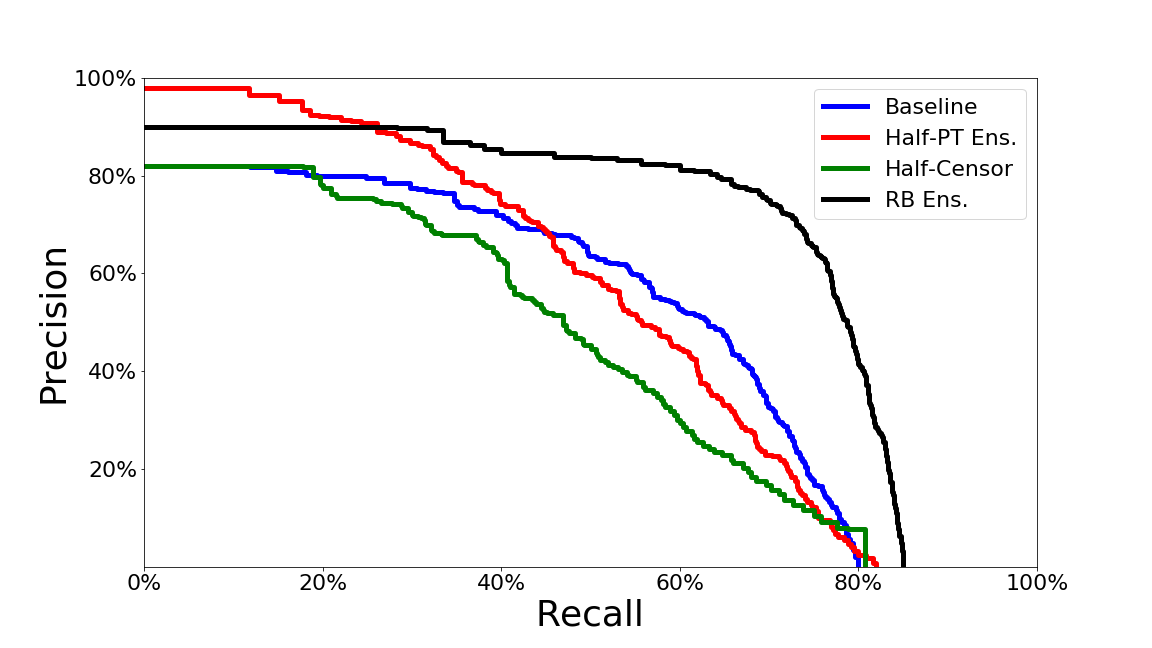}
	\caption{\textbf{PR Curves for Different Models.}}
	\label{fig:roc}
\end{figure}

We compare the aforementioned ensembling techniques (Table \ref{table:ensembling}) and visualize the PR curves for four of these networks (Fig. \ref{fig:roc}).  The colors of the PR curves of Fig. \ref{fig:roc} are shown next to the names of the models in Table \ref{table:ensembling}.

\begin{table}[htb!]
 \caption{\textbf{Performance of Different Ensembling Techniques}}\label{table:ensembling}
\begin{center}
\scriptsize
\begin{tabular}{L{2cm}C{1.5cm}C{2.2cm}C{2.3cm}C{2.9cm}}
\toprule[1.5pt]
Network & \# of Nets & mAP (95\% CIs) & Sens. at 80\% Prec. & TP DICE at 80\% Prec.\\
\midrule
\midrule
DeepLabv3 \hfill \crule[blue]{3mm}{3mm} & 1 & 46 (44,47) & 22 & 72 \\
\midrule
DeepLabv3 & 3 & 47 (45,48) & 23 & 72 \\
DeepLabv3 & 10 & 47 (46,48) & 24 & 73 \\
DeepLabv3 & 30 & 47 (46,48) & 24 & 73 \\
\midrule
Original FCN & 1 & 32 (29,33) & 15 & 65\\
U-Net & 1 & 39 (36,42) & 18 & 69\\
SegNet & 1 & 38 (36,41) & 19 & 69\\
FPN & 1 & 44 (42,46) & 21 & 71\\
PSP Net & 1 & 46 (44,48) & 21 & 72\\
Architecture Ensemble & 6 & 43 (40,46) & 20 & 71\\
\midrule
\nicefrac12 Patients & 1 & 44 (42,46) & 20 & 70\\
\nicefrac12 Patients & 3 & 48 (46,49) & 26 & 72\\
\nicefrac12 Patients & 10 & 51 (49,52) & 30 & 73\\
\nicefrac12 Patients \hfill \crule[red]{3mm}{3mm} & 30 & 58 (57,59) & 34 & 74\\
\midrule
\nicefrac12 Censored \hfill \crule[green]{3mm}{3mm} & 1 & 42 (40,44) & 21 & 73\\
\nicefrac12 Censored & 3 & 49 (47,51) & 35 & 74\\
\nicefrac12 Censored & 10 & 55 (54,57) & 48 & 73\\
\nicefrac12 Censored \hfill \crule[black]{3mm}{3mm}& 30 & \textbf{64 (63,65)} & \textbf{67} & \textbf{75}\\
\end{tabular}
\end{center}
\end{table}
\normalsize

With different hyperparameter/initialization ensembling, the performance difference w.r.t.\ the baseline single DeepLabv3 model is not statistically significant.  For architecture ensembling, performance actually decreases.  Ensembling networks trained on different patient cohorts improves mAP, but the minimal gains in sensitivity (and Fig. \ref{fig:roc}) show that these gains result from increasing precision rather than recall.  In comparison, the RB method (censoring lesions rather than patients) shows a statistically significant increase in mAP values through higher sensitivity (Table \ref{table:ensembling} and Fig. \ref{fig:roc}).

\begin{figure}[htb!]
	\centering
	\includegraphics[width=0.95\textwidth]{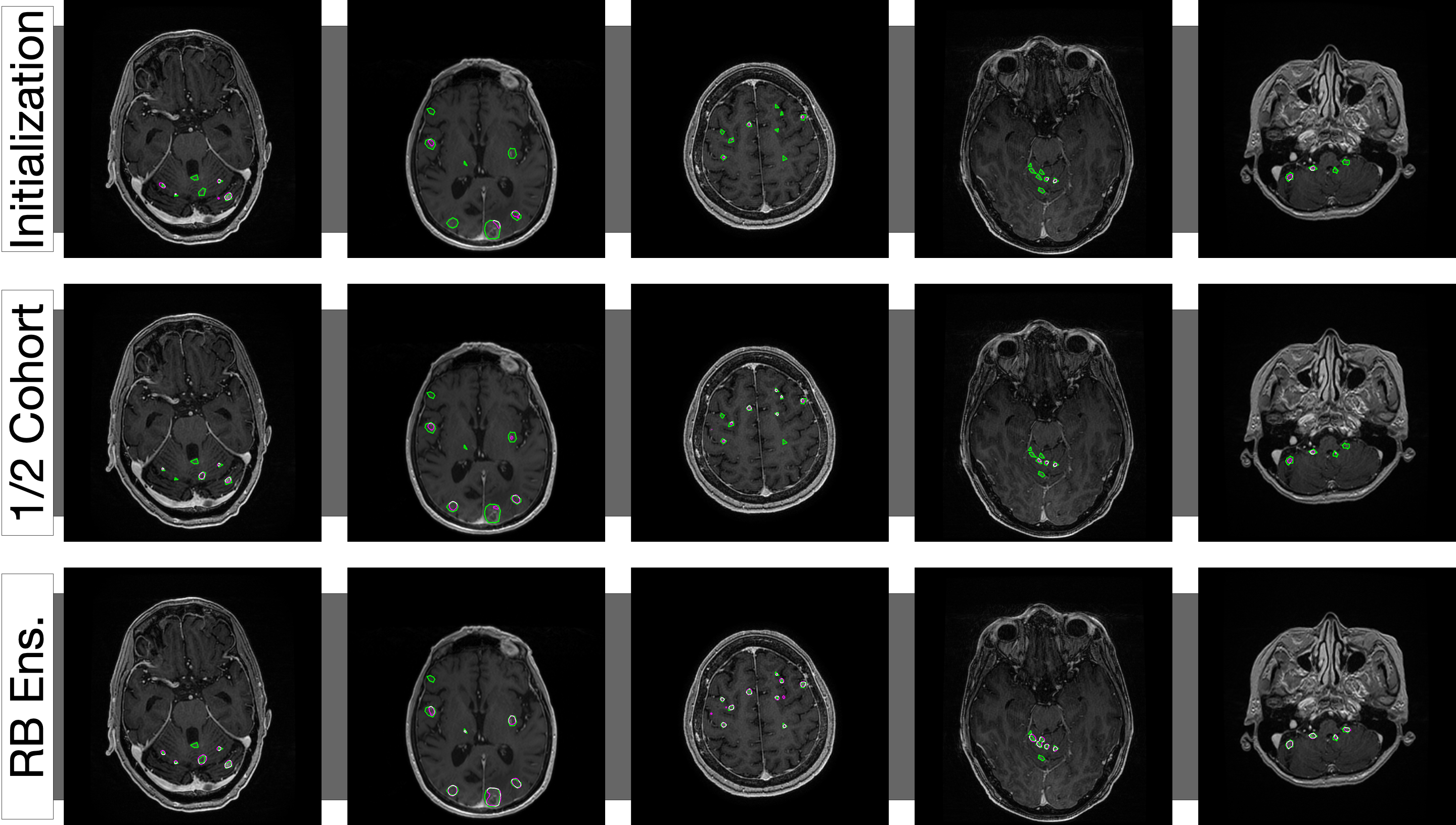}
	\caption{\textbf{Visualization of Different Ensembles.}  \small{We show the five MR slices from the test set with the most lesions.  Expert ``target'' annotations are outlined in green.  The predictions' binarized CC's boundaries are shown in magenta.  Overlap is white.}}
	\label{fig:examples}
\end{figure}

A qualitative comparison of three ensembling techniques (1. random initializations/hyperparameters, 2. randomly sampling half the patient cohorts, and 3. the RB ensemble), each ensembling 30 networks, can be seen in Fig. \ref{fig:examples}.  Frames with larger numbers of lesions were chosen to better demonstrate the differences in sensitivity between the techniques.  The RB ensemble has the highest sensitivity, followed by the ensemble of networks trained on different patient cohorts, and finally the ensemble of different initializations/hyperparameters.

%\subsection{Wide Res-Net Study}

In Table \ref{table:wideresnet}, we explore ensembling different numbers of a $k=\nicefrac12$ WRN, denoted WRN(\nicefrac12), and compare it to a single $k=2$ WRN, denoted WRN(2), trained on the full data.  With the same amount of compute ($1\times$ WRN(2) vs. $16\times$ WRN(\nicefrac12)), using the RB ensemble framework gives statistically significant improvement in results compared with investing all the compute into a single network.

\vspace{-10mm}
\begin{table}[htb!]
\caption{\textbf{Wide Res-Net Study}}\label{table:wideresnet}
\begin{center}
\scriptsize
\begin{tabular}{L{2.3cm}C{1.5cm}C{2.2cm}C{2.3cm}C{2.9cm}}
\toprule[1.5pt]
Network & \# of Nets & mAP (95\% CIs) & Sens. at 80\% Prec. & TP DICE at 80\% Prec.\\
\midrule
\midrule
WRN(2) on Full & 1 & 35 (32,38) & 18 & 68\\
\midrule
& 1 & 26 (22,29) & 15 & 70\\
\multirow{2}{*}{WRN(\nicefrac12) on}& 2 & 33 (30,35) & 22 & 70\\
\multirow{2}{*}{\nicefrac12 Censored}& 4 & 37 (34,40) & 26 & 72\\
& 8 & 40 (38,42) & 31 & 72\\
& 16 & \textbf{43 (41,45)} & 35 & 72
\end{tabular}
\end{center}
\end{table}
\vspace{-15mm}

\section{Discussion}

\textbf{Ensembling without enough statistical variance between networks yields little benefit.}  Ensembling 30 DeepLabv3's gives statistically insignificant improvement in all metrics tested.  As they are essentially the same networks trained with the same dataset, they predict very similar outputs, with minimal benefit from ensembling.  Ensembling different architectures does increase variance, but worse-performing networks may simply predict a subset of correct predictions of better ones, consistent with the performance loss seen.  Ensembling networks trained on a random half of the patients gives statistically significant improvement, but the gains in mAP are due to precision, not recall (Figs. \ref{fig:roc} and \ref{fig:examples}).

\textbf{Ensembled networks need to have more sensitivity.}  Knowing that incorporating lopsided boostrap loss (Eq. \ref{eq:lopsided}) makes networks more sensitive \cite{yi2020brain}, we induce stochasticity through randomizing lesions rather than patients.  Individual networks become much more sensitive and perform worse than a single ``correctly'' trained DeepLabv3 network, but the ensembled true positive predictions from our network are more consistent than the false positive predictions.  Thus, we average out errors while maximally preserving correct predictions.

\textbf{WRN experiments suggest that ensembling several weaker networks can be much stronger than using a single strong network.}  A key limitation of ensembling networks in general is the high computational cost; compute increases linearly with networks in an ensemble, but performance increases logarithmically at best.  The results indicate that creating parallelized ensembling can be a better use of compute than simply increasing a single network's compute.

\textbf{Our study has known limitations.} While we took care to optimize the baseline model to allow rigorous performance comparisons, minimizing hyperparameter search for other models means that they might still be improved, making the validity of those comparisons less clear.  Furthermore, using WRNs allowed us to equalize compute, but they were not created for segmentation, as seen by much lower performance compared to the baseline DeepLabv3.  An important future direction will be to create an ensemble of network architectures with total compute equal to that of a single DeepLabv3 and train them with the RB protocol.  Finally, RB ensembling is developed for tasks with many localized lesions, such as brain metastases; it may be less effective in segmenting lesions that are large, connected, and limited per patient (e.g. glioblastoma).

\section{Conclusion}

In our work, we outline a new ensembling technique we call ``Random Bundle.''  By using the lopsided bootstrap loss developed to train on segmentation annotations with a high false negative rate, we proceed to train each sub-network in our ensemble on our data with a random 50\% subset of lesions censored out.  This result achieves the highest mAP, sensitivity at 80\% precision, and true positive DICE coefficient out of all of the different ensemble techniques we tested.  In addition, by creating ensembles with WRNs of different width parameters, we show that RB ensembling can be a more effective use of compute.  Such advances could elevate more segmentation tasks to a clinically relevant stage without exponentially scaling data collection.

\section*{Acknowledgments}

%Censored for double-blind review.  To understand page length, we create fake text in this area to model how long our Acknowledgements section will actually be.
Data provided by Stanford Hospital.  We acknowledge the following grants: T15 LM 007033, U01CA142555, 1U01CA190214, 1U01CA187947, and U01CA242879.

\bibliographystyle{splncs04}
\bibliography{references}
\end{document}